\RequirePackage{xcolor}

\def\ispreprint{1}

\ifx\ispreprint\undefined
    
\else
    \documentclass[twocolumn,10pt]{article}
    \usepackage[margin=1.25cm,right=0.75cm,top=.5cm,bottom=1.75cm]{geometry}

    \usepackage{cite}
    \usepackage{amsmath,amssymb,amsfonts}
    \usepackage{algorithmic}
    \usepackage{graphicx}
    \usepackage{textcomp}
    \usepackage{adjustbox}
    \usepackage{fancyvrb}
    \usepackage{authblk}
    \usepackage{hyperref}

    \usepackage[font=small,labelfont=bf,skip=0.25cm]{caption}
    \renewcommand{\vec}[1]{\mathbf{#1}} 
    \newcommand{\Rr}{\mathbb R}
    
    \usepackage{tikz}
    
    \newcommand\submittedtext{%
      \footnotesize This work has been submitted to the IEEE for possible publication. Copyright may be transferred without notice, after which this version may no longer be accessible.}
    
    \newcommand\submittednotice{%
    \begin{tikzpicture}[remember picture,overlay]
    \node[anchor=south,yshift=10pt,fill=white] at (current page.south) {\fbox{\parbox{\dimexpr0.65\textwidth-\fboxsep-\fboxrule\relax}{\submittedtext}}};
    \end{tikzpicture}%
    }

\begin{document}
        \title{PatchMorph: A Stochastic Deep Learning Approach for Unsupervised 3D Brain Image Registration with Small Patches}

    \author[1,*]{Henrik Skibbe}
    \author[1,2]{Michal Byra}
    \author[3]{Akiya Watakabe}
    \author[3,4]{Tetsuo Yamamori}
    \author[5]{Marco Reisert}
    \affil[1]{\small Brain Image Analysis Unit, RIKEN Center for Brain Science, Wako, Saitama, Japan.}
    \affil[2]{\small Institute of Fundamental Technological Research, Polish Academy of Sciences, Warsaw, Poland.}
    \affil[3]{\small Laboratory of Haptic Perception and Cognitive Physiology,
    RIKEN Center for Brain Science, Wako, Saitama, Japan.}
    \affil[4]{\small Department of Marmoset Biology and Medicine, Central Institute for Experimental Animals, Kawasaki, Japan.}
    \affil[5]{\small Department of Radiology, Medical Physics, and Department of Stereotactic and Functional Neurosurgery, Medical Center Freiburg, Faculty of Medicine, University of Freiburg, Freiburg, Germany.}
    \affil[*]{contact:\textit{henrik.skibbe@riken.jp}}
\fi

\maketitle

\ifx\ispreprint\undefined
\else
    \thispagestyle{empty}
\fi

\begin{abstract}
We introduce "PatchMorph," an new stochastic deep learning algorithm tailored for unsupervised 3D brain image registration. Unlike other methods, our method uses compact patches of a constant small size to derive solutions that can combine global transformations with local deformations. This approach minimizes the memory footprint of the GPU during training, but also enables us to operate on numerous amounts of randomly overlapping small patches during inference to mitigate image and patch boundary problems. PatchMorph adeptly handles world coordinate transformations between two input images, accommodating variances in attributes such as spacing, array sizes, and orientations. The spatial resolution of patches transitions from coarse to fine, addressing both global and local attributes essential for aligning the images. Each patch offers a unique perspective, together converging towards a comprehensive solution. Experiments on human T1 MRI brain images and marmoset brain images from serial 2-photon tomography affirm PatchMorph's superior performance.
\end{abstract}

\ifx\ispreprint\undefined

\else
     \submittednotice
\fi

\section{Introduction}\label{sec:introduction}

Image registration holds a central position in medical and neuroscientific research and applications. By spatially aligning one image with another, it facilitates quantitative comparisons of features among different subjects or across distinct time points. This alignment becomes especially crucial for group studies, where, for instance, brain magnetic resonance imaging (MRI) scans must be aligned to a reference template.

The task of image registration is fraught with challenges. Images obtained by techniques such as MRI or microscopy often contain artifacts and may be constrained in terms of spatial resolution. In addition, inherent differences within imaged subjects, such as brain scans of separate individuals, can make establishing an exact one-to-one mapping difficult, if not impossible.

Balancing these challenges, image registration seeks to spatially modify an image to closely resemble another while preserving medically or scientifically important features. 
Classical tools, such as ANTs \cite{avants2011reproducible} and Elastix \cite{klein2009elastix}, consider registration as an iterative optimization problem. The overarching goal is the alignment of content from a "moving" image to best match that in a "fixed" image. 

Metrics like mean-square difference or normalized cross-correlation are employed to measure this similarity. Furthermore, to retain image topology, data priors are commonly used to ensure topology remains consistent between adjacent voxels.

Recent advances in deep learning and the corresponding hardware developments have enabled architectures that mimic these classical techniques \cite{haskins2020deep}. Specifically, VoxelMorph \cite{dalca2018unsupervised} and its variants \cite{balakrishnan2019voxelmorph,hoffmann2021synthmorph} have shown that deep neural networks can learn in an unsupervised way to predict intricate warp fields between two input images, offering an advantage in terms of time over iterative optimizations seen in methods such as ANTs or Elastix. While deep learning algorithms have also facilitated supervised training techniques using labels \cite{hu2018weakly} or synthetically generated data \cite{eppenhof2018pulmonary,hoffmann2021synthmorph}, our work focuses on unsupervised training, maximizing a similarity metric \cite{guo2019multi,balakrishnan2019voxelmorph} between the moving and fixed images, akin to ANTs or Elastix. Extensions of VoxelMorph, which employ cascades of the input images at multiple scales, akin to ANTs, address the registration task with a coarse-to-fine approach \cite{mok2020large,zhang2021cascaded}. Yet, these deep learning techniques often have their own set of limitations, notably the necessity for paired input images to have congruent array sizes. Another constraint is that most of these approaches feed entire images into networks, which poses a limitation due to large GPU memory requirements during training.

Emerging techniques such as implicit neural representations (INR) for image registration \cite{wolterink2022implicit,byra2023exploring} combine artificial neural network strategies with classical optimization approaches. These methods take coordinates as inputs and can therefore handle image pairs with different array sizes or voxel resolutions. They have shown better registration outcomes over techniques such as VoxelMorph and even outperformed ANTs in certain respects. However, like traditional methods, they require optimization for each new pair of images, thereby not benefiting from the rapid inference times associated with deep learning methods.

"PatchMorph," our novel architecture, employs a Deep Neural Patchwork approach \cite{reisert2022deep} for image registration, operating in a coarse-to-fine manner. A Deep Neural Patchwork employs a cascading network architecture for semantic segmentation of small patches in convolutional neural networks (CNNs), efficiently managing memory constraints while preserving global context in biomedical imaging. PatchMorph uses this approach to sequentially refine the focus from  coarsely sampled patches with a large field of view to smaller, detailed patches, leveraging stacked CNNs that cascade information to improve registration accuracy without exceeding compact patch sizes.

Solving the problem in a coarse-to-fine manner is not new. Coarser scales are often used for a rough alignment, and finer resolutions to optimize details. It is a fundamental part of a registration pipeline in ANTs, but also a key component in recent deep learning approaches such as \cite{cetin2022multi,mok2020large,zhang2021cascaded,chatterjee2023micdir}. While ANTs and approaches like \cite{mok2020large,zhang2021cascaded} operate on the entire images in different scales, estimating and refining the transformation in a coarse-to-fine manner, other methods take a different approach. For instance, \cite{cetin2022multi}, processes centered patches in two different resolutions, original and half, from both the moving and fixed images, oncatenating and feeding them into neural networks to predict the warping field. Similary, \cite{chatterjee2023micdir} uses both the original and downscaled versions of the fixed image within the same network. These last two approaches share the concept of adding global context to the network, albeit in slightly distinct ways.

PatchMorph estimates and refines, similar to ANTs, a transformation that comprises global and local deformations in a coarse-to-fine manner. But distinctively, PatchMorph retains a consistent, but relatively small, patch size; a typical size being $32^3$ voxels across all scales. Similar to ANTs, but in contrast to most deep learning alternatives such as VoxelMorph, PatchMorph can process moving and fixed images with varied array sizes, orientations, and spatial resolutions. This capability eliminates the need for operations like rescaling or padding to make sure that the sizes of the arrays of the moving and fixed images are the same, which conserves valuable GPU memory. 

PatchMorph selects large numbers of pairs of patches from random positions within both fixed and moving images, successively estimating and refining the displacement field between the two. On broader scales, we employ coarse transformations to align images, focusing on position, scale, and global orientation. On finer scales, our approach delves into the nuances. 

Importantly, during training, patches on finer scales consistently encapsulate details inherent to their corresponding coarser-scale patches. The PatchMorph architecture is designed to understand the affine transformations that map results from a patch on one scale to its more detailed successor. This capability enables seamless propagation of updates across scales. Crucially, this allows for the optimization of a global registration problem using small patches across all scales, thereby eliminating the need to compute the global displacement field for the entire input images during training, a significant advantage over VoxelMorph.

In the inference phase, PatchMorph estimates the global displacement field for the entire image. PatchMorph randomly selects thousands of pairs of patches, estimating the displacement between these patch pairs, and gradually coalesces the estimations into an expansive update for the global displacement field. This method then undergoes refinement in a stepwise, coarse-to-fine approach, mirroring the multi-scale optimization employed during training.  The results from randomly overlapping patches are averaged, eliminating interpolation artifacts. 

In experiments, PatchMorph demonstrates superior performance compared to other methods, all while maintaining the integrity of the deformation field.

\section{Method}

\subsection{Preliminaries}
Let $\vec I:\Rr^3 \to \Rr$ be a 3D image stack. With $\vec X:[0,1]^3 \to \Rr^3$ we denote a coordinate field defined on a normalized grid, and with $\vec d=\vec X_\text{m} - \vec X_\text{f}$ the dense displacement field (DDF) with respect to two coordinate fields. With $\vec r\in\Rr^3$, we denote a coordinate vector. Our objective is, given a fixed image $\vec I_\text{f}$ and a moving image $\vec I_\text{m}$, to seek the global DDF to compute the coordinate field $\vec X_\text{m}$ that fulfills:
\begin{align}
\vec I_\text{f}(\vec r)\simeq\vec I_\text{m}(\vec X_\text{m}(\vec r)). 
\label{eq:imgreg}
\end{align}
With the symbols $\vec T$ and $\vec A$ we denote 3D affine $4\times4$ matrices with 12 degrees of freedom, where the last row is the constant unit vector.

\subsubsection{Array and World Coordinates}
\begin{figure}
\includegraphics[width = \columnwidth]{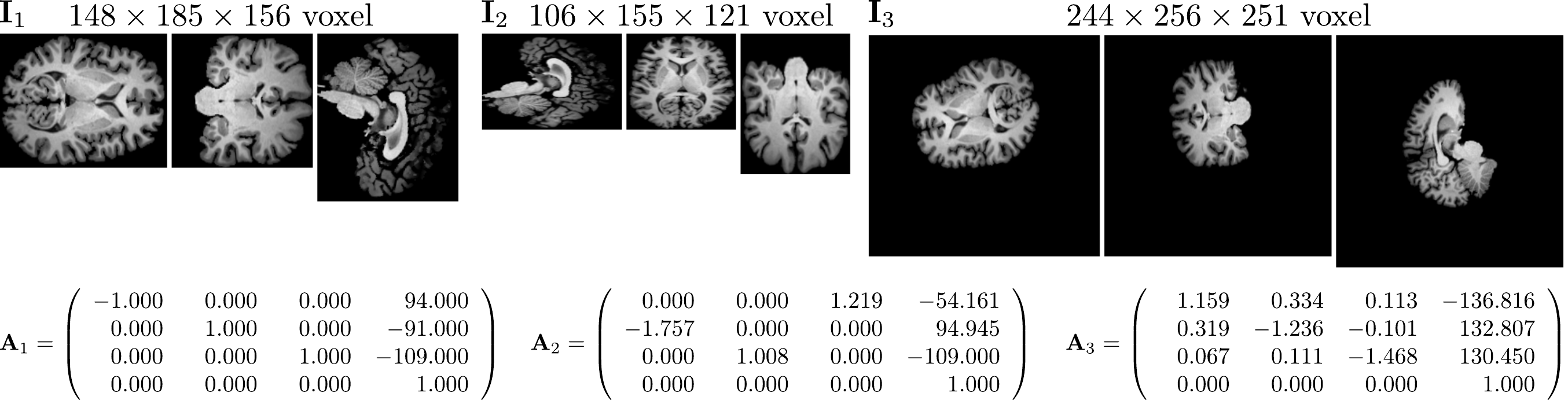}
\caption{
This figure displays orthogonal sections from three image stacks $\vec I_i$, each paired with a world transformation matrix that maps image coordinates to physical world coordinates. PatchMorph utilizes these matrices to accurately sample data from the original arrays. Notably, despite their varied appearances, all arrays in this example depict the same physical image.
}
\label{fig:nifti}
\end{figure}

MRI and microscopy image stacks, $\vec I_i$, stored in formats such as Nifti and Tiff, are accompanied by affine 3D matrices $\vec A_i$ that map array coordinates to physical world coordinates. We denote the inverse of this matrix as $\textbf{T}_i:=(\vec A_i)^{-1}$. Figure \ref{fig:nifti} illustrates how different arrays and orientations can represent the same image through such transformations, as expressed by $\vec {I}_1(\vec{T}_1\vec{r})=\vec{I}_2(\vec{T}_2\vec{r})=\vec{I}_3(\vec{T}_3\vec{r})$.

PatchMorph leverages this by directly sampling patch data from original arrays without the need for padding or resolution alteration, thus maintaining the intended resolution, orientation, and scale. It simplifies the use of fixed ($\textbf{I}_\text{f}$, $\textbf{T}_\text{f}$) and moving ($\textbf{I}_\text{m}$, $\textbf{T}_\text{m}$) image stacks. The affine transformation matrices, $\textbf{T}_\text{i}$, integrate rotation, scaling, and translation, providing 12 degrees of freedom.

\subsection{PatchMorph}

The fundamental input for PatchMorph consists of two images: a fixed image, denoted $\vec{I}_\text{f}$, and a moving image, $\vec{I}_\text{m}$. Each of these images is paired with its respective world transformation matrices $\textbf{T}_\text{f}$ and $\textbf{T}_\text{m}$. From these images we draw the patches.

In the trivial case where the array sizes of $\vec{I}_\text{f}$ and $\vec{I}_\text{m}$ are identical, and both $\textbf{T}_\text{f}$ and $\textbf{T}_\text{m}$ are identical diagonal matrices, we could just determine patch coordinates with respect to one of these arrays, and gather the image content from both images. But in general, the sizes,  spacing and orientation of the image arrays may differ.

\begin{figure}
\centering
\includegraphics[width = 1.0\columnwidth]{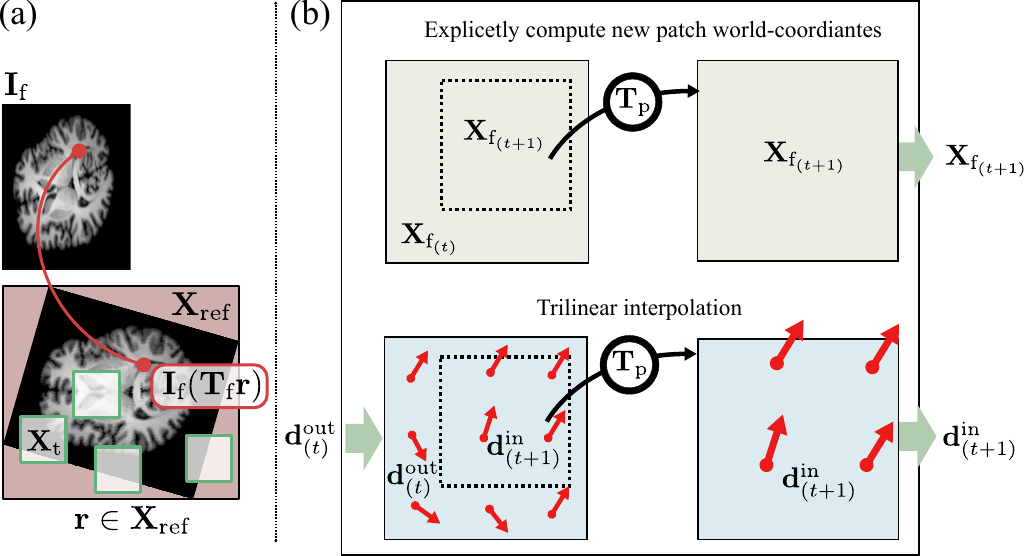}
\caption{
Panel (a): PatchMorph selects patch coordinates $\vec X_\text{t}$ from the 'working canvas' world coordinate field $\vec X_\text{ref}$. This canvas is determined by the axis-aligned bounding box enclosing the fixed image $\vec I_\text{f}$ in physical space. Panel (b): For each iteration at $t>0$, the patch coordinates for the moving image are updated using the resulting displacement field $\vec d_{\text{out}{(t)}}$ in combination with the affine patch coordinate transformation $\vec T_{\text{ p}_{(t+1)}}$, setting up the input for the subsequent scale.
}
\label{fig:spatching}
\end{figure}

Therefore, PatchMorph operates on a coordinate field we have named the "working canvas", represented as $\vec X_\text{ref}$ defined by an affine matrix $\vec T_\text{ref}$. This canvas serves as tight, axis-aligned bounding box enclosing the coordinate space of the image $\vec{I}_\text{f}$, ensuring that for every array coordinate vector $\vec r$ from the domain of $\vec{I}_\text{f}$, the transformed coordinate vector, $(\vec{T}_\text{f})^{-1}\vec r$, resides within $\vec X_\text{ref}$. See panel (a) in Fig. \ref{fig:spatching} for an illustration.

\begin{figure}
\includegraphics[width = \columnwidth]{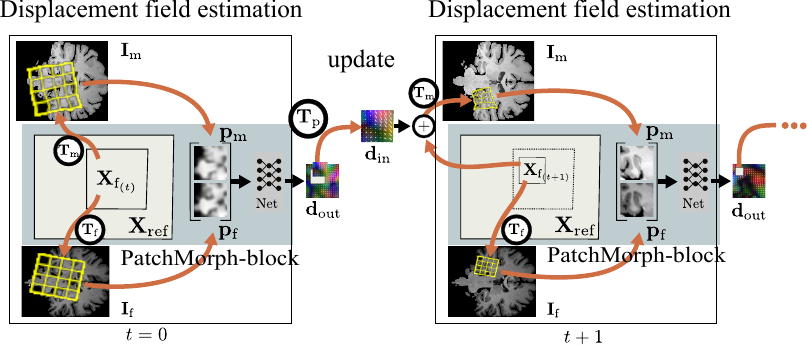}
\caption{PatchMorph orchestrates a sequence of PatchMorph-blocks from coarse to fine scales. Within each block, it samples small 3D patches from both fixed and moving images, standardizes their spatial resolution, and employs a CNN to estimate and refine their relative displacement.
}
\label{fig:teaser}
\end{figure}

\begin{figure}
\includegraphics[width = \columnwidth]{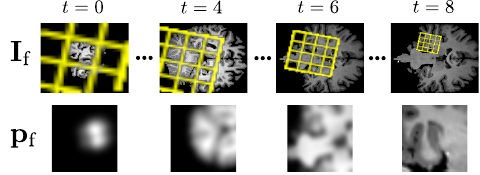}
\caption{
This figure displays cross-sections of a 3D image stack, $\mathbf{I}_{\text{f}}$, from which 3D patches $\vec p_{\text{f}}$ are extracted. Patches are sampled using a coarse-to-fine strategy, ensuring that each finer-resolution patch is contained within its coarser-scale predecessor. The bottom row shows cross sections of the patches. The array size of each patch is consistently maintained across all scales.
}
\label{fig:imgarray}
\end{figure}

Fig. \ref{fig:teaser} shows a simplified sketch of the PatchMorph architecture. PatchMorph is starting on a coarse scale (represented by scale $t=0$) and becomes progressively finer. Each scale deploys a PatchMorph-Block. At each scale, PatchMorph extracts patches of a consistent array size, but with progressively increasing isotropic spatial resolution; see Fig. \ref{fig:imgarray} for an example.

At $t=0$, PatchMorph randomly creates patches of world coordinate fields $\vec X_{\text{f}_{0}}$ extracted from $\vec X_\text{ref}$ given affine matrices $\vec T_{\text{p}_0}$. It also initializes the patch coordinate field for the moving image to be $\vec X_{\text{m}_{0}}=\vec X_{\text{f}_{0}}$. PatchMorph then extracts the corresponding image patches $\vec p_\text{f},\vec p_\text{m}$ from both fixed and moving images using linear interpolation. 

After extraction, patches have uniform size and resolution. They are then concatenated along the feature dimension and input into an image registration CNN. The specifics of this process will be explained later in the context of the PatchMorph-Block and our CNN architectures. The output from the CNN is a three-channel image, $\vec d_{\text{out}_{0}}$, which represents the estimated DDF. This field refines the moving coordinate field according to $\vec X_{\text{f}_{0}}+\vec d_{\text{out}_{0}}$.

As visualized in Fig. \ref{fig:imgarray}, each subsequent iteration over $t$ extracts new patches from the fixed image that progressively increases the resolution and narrow the focus. During subsequent iterations, PatchMorph calculates a new affine transformation $\vec T_{\text{p}_{(t+1)}}$. This affine transformation serves to map the current patch coordinates for the fixed image patch, $\vec X_{\text{f}_{(t)}}$, onto a patch of new coordinates, $\vec X_{\text{f}_{(t+1)}}$. The newly derived coordinates essentially provide the coordinates for a magnified view of a sub-patch from the prior patch, where $\vec X_{\text{f}_{(t+1)}}(\vec r):=\vec X_{\text{f}_{(t)}}(\vec T_{\text{p}_{(t+1)}}\vec r)$ ; see the top row of panel (b) in Fig. \ref{fig:spatching}. 

For $t>0$, the patch coordinates for the moving patch $\vec X_{\text{m}_{(t+1)}}$ at $t+1$ are updated based on inputs $\vec d_{\text{in}_{(t+1)}}$. To transition from the output $\vec d_{\text{out}_{(t)}}$ to the input for the next iteration $\vec d_{\text{in}_{(t+1)}}$, we employ the patch coordinate mapping $\vec T_{\text{p}_{(t+1)}}$ along with linear interpolation yielding  $\vec X_{\text{m}_{(t+1)}}:=\vec X_{\text{f}_{(t+1)}}+\vec d_{\text{in}_{(t+1)}}$; see bottom row of panel (b) in Fig. \ref{fig:spatching}.

It is important to note that in our implementation, the initial transformation $\vec T_{\text{p}_{0}}$ involves all 12 degrees of freedom, including rotation, shift, and scaling, while for $t>0$, the transformation strictly involves operations of (up)-scaling and shifting. 

This relationship is important during training. By establishing this iterative relationship between the coarser and more detailed patch representations, the DDF estimations from earlier iterations are leveraged to update and improve the coordinate field of the moving image patch. Consequently, with richer content details and updated positions, the CNN is redeployed to attain more refined results. In this way, PatchMorph can optimize a global registration problem on small patches without the need to compute the global displacement field for the entire input images.

\subsubsection{PatchMorph-Block}

\begin{figure*}
\includegraphics[width = \textwidth]{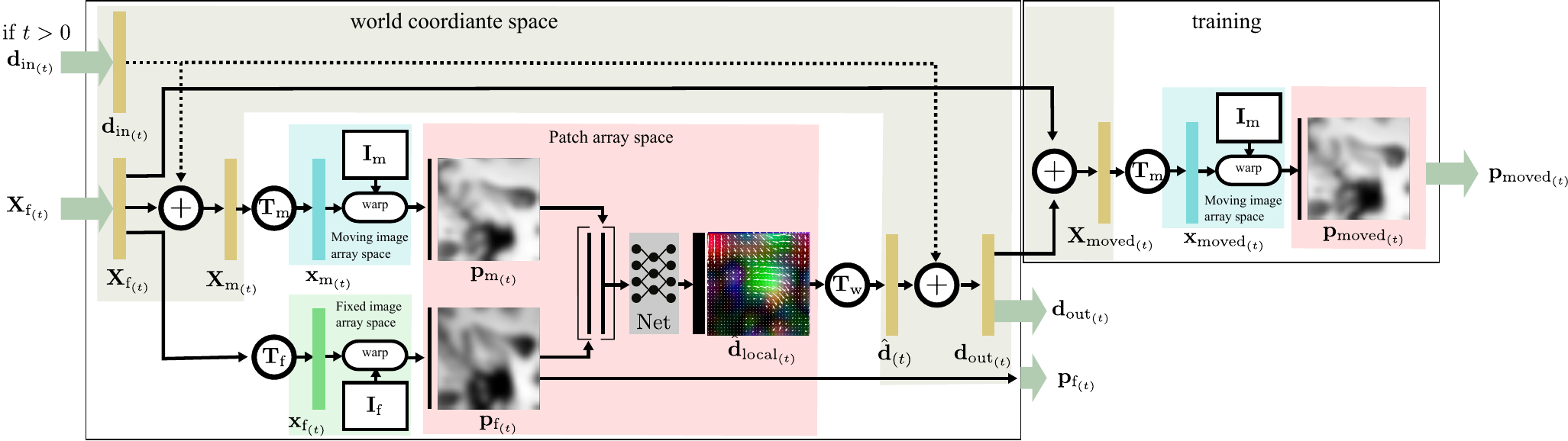}
\caption{
This figure delineates the PatchMorph-Block architecture. It details how the block processes inputs, world coordinates for the fixed image patch $\vec X_{\text{f}{(t)}}$ and the relative displacement patch $\vec d_{\text{in}{(t)}}$ from the previous scale ($t>0$), to output the updated relative displacement $\vec d_{\text{out}{(t)}}$. The block computes world coordinates for the moving patch, extracts and concatenates image patches from the image stacks, and applies a CNN to estimate the local voxel displacement field. The resulting field is then translated back to world coordinates using an affine transformation, combined with the input from the previous iteration to form the final output displacement field $\vec d_{\text{out}_{(t)}}$. 
}
\label{fig:netarch}
\end{figure*}

Figure \ref{fig:netarch} provides a detailed representation of the PatchMorph-Block architecture. It accepts two input arguments: the world coordinates $\vec X_{\text{f}_{(t)}}$ for the current fixed image patch and, when $t>0$, a relative displacement patch $\vec d_{\text{in}_{(t)}}$ from the previous iteration. The primary output, which is utilized during both training and inference, is the updated relative displacement vector patch $\vec d_{\text{out}_{(t)}}$.

Given the inputs, PatchMorph determines the world coordinates of the current moving patch using $\vec X_{\text{m}_{(t)}}:=\vec X_{\text{f}_{(t)}}+\vec d_{\text{in}_{(t)}}$. For $t=0$, $\vec X_{\text{m}_{(t)}} = \vec X_{\text{f}_{(t)}}$. 
Based on both, the transformation matrices $\vec T_\text{f}$ and $\vec T_\text{m}$, we can map the world coordinate fields to  array coordinate fields $\vec x_{\text{f}_{(t)}},\vec x_{\text{f}_{(m)}}$. Based on them, we extract the corresponding image patches $\vec p_{\text{f}_{(t)}}$ and $\vec p_{\text{m}_{(t)}}$ from the image stacks $\vec I_\text{f}$ and $\vec I_\text{m}$. Post-extraction, both patches are concatenated and fed into a CNN that estimates the relative voxel displacement field $\hat{\vec d}_{\text{local}_{(t)}}$ between the two image patches. Using the affine matrix $\vec T_{\text{w}_{(t)}}$, we translate the update of the displacement field back to the world coordinates, represented as $\hat{\vec d}_{(t)}$. The matrix $\vec T_{\text{w}_{(t)}}$ can be computed from the patch array size, the matrix $\vec T_\text{ref}$,  and the matrices $\vec T_{\text{p}_{t=\{0,\cdots,t\}}}$. If available, we update these coordinates with $\vec d_{\text{in}_{(t)}}$, deriving the refreshed displacement field 
\begin{align}
\vec d_{\text{out}_{(t)}}:=\underset{=\hat{\vec d}_{(t)}}{\underbrace{\vec T_{\text{w}_{(t)}}~\hat{\vec d}_{\text{local}_{(t)}}}} +\begin{cases}
\vec d_{\text{in}_{(t)}} ~\text{if $t>0$}\\
\vec 0 ~\text{else}.
\end{cases}
\end{align}

\subsubsection{Training}
During training, we cropped zero-boundaries of the images to minimize the array size, and we modify the translation vectors of $\vec T_\text{f}$ and $\vec T_\text{m}$ so that the world coordinate center of the images aligns with the origin. We also added the left-right mirrored copy of each image to the training set.

For each $t$, we update the world coordinates of the moving patch $\vec X_{\text{moved}_{(t)}}:=\vec X_{\text{f}_{(t)}}+\vec d_{\text{out}_{(t)}}$ to obtain an updated version of the moving image patch denoted as $\vec p_{\text{moved}_{(t)}}$. The training objective is to optimize the similarity between the outputs, $\vec p_{\text{moved}_{(t)}}$, and the fixed image patches, $\vec p_{\text{f}_{(t)}}$ across all scales in an unsupervised way using an image similarity loss such as the normalized cross correlation (NCC).

After the training process, equation $\vec X_{\text{moved}_{(t)}} := \vec X_{\text{f}_{(t)}} + \vec d_{\text{out}_{(t)}}$ determines the world coordinates of the image patch in the moving image that optimally align with the fixed image patch situated at location $\vec X_{\text{f}_{(t)}}$ within the fixed image.

\begin{figure}
\includegraphics[width = \columnwidth]{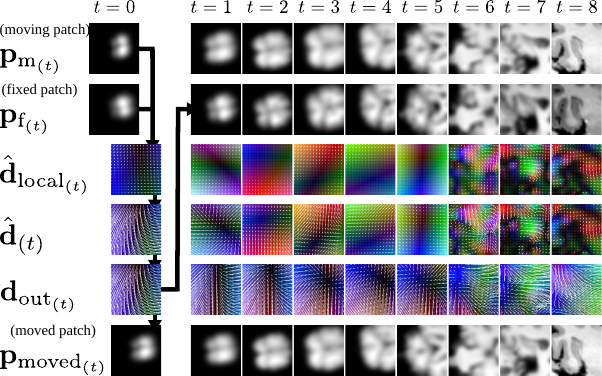}
\caption{
This figure illustrates cross-sections from a PatchMorph-block's inputs and outputs. The top row displays the moving and fixed image patches, while the bottom row shows the moved image patches across all nine scales. The primary objective is to compute the dense displacement field $\vec d_\text{out}$, aligning the moving patch content to optimally match the fixed patch.
}
\label{fig:patching}
\end{figure}

Figure \ref{fig:patching} displays examples of fixed, moving, and moved image patches after training, extracted based on coordinate patches in nine different scales.

\subsubsection{Inference}
During inference, we adjust the translation vector of $\vec T_\text{m}$ to align the center of the moving image's world coordinates with that of the fixed image, correcting this shift afterward in post-processing.

The goal is to construct the global DDF $\vec D$, satisfying the image registration equation \eqref{eq:imgreg}, where $\vec X_\text{m}=\vec X_\text{ref}+\vec D$.

Unlike training, inference does not require a cascade of nested patches. To assemble the global $\vec D$, PatchMorph generates successive updates $\vec D_{(t)}$ at each scale, culminating in: $\vec D := \sum_t \vec D_{(t)}$.

\begin{figure}
\includegraphics[width = \columnwidth]{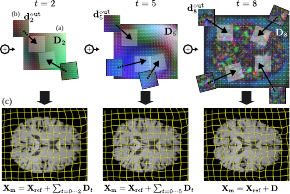}
\caption{
The goal during inference is to develop a global dense displacement field $\vec D$, refined iteratively from coarse to fine scales. As shown in (a), for each scale, an update $\vec D_{(t)}$ is estimated by combining displacement estimates from numerous patches $\vec d_{\text{out}_t}$ (illustrated in (b)). With each new iteration $t+1$, this refined field guides the selection of patch pairs with increased resolution for more accurate updates. The number of patches and the resolution of the displacement field increase proportionally with the patch resolution (further details in the main text). Panel (c) depicts the deformations of the updated coordinate field at iteration $t$, demonstrating the alignment of the moving image to the fixed image.
}
\label{fig:inference01}
\end{figure}

Initially, PatchMorph creates a low-resolution displacement field $\vec D_{(0)}$ for the first scale. As the process advances to subsequent iterations, the spatial resolution progressively increases, with each iteration yielding a global update $\vec D_{(t)}$, as visualized in Fig. \ref{fig:inference01}. The array resolution of each $\vec D_{(t)}$ is enhanced to match the increased spatial resolution of the patches. To accommodate the increase in resolution from $\vec D_{(t)}$ to $\vec D_{(t+1)}$, linear interpolation is utilized to upscale the fields with lower array resolutions.

At each iteration $t$, PatchMorph samples a large number of coordinate patches $\vec X_{\text{f}_{t}}$ from the canvas $\vec X_\text{ref}$. For $t>0$, displacement updates $\vec d_{\text{in}_{(t)}}$ are interpolated from $\vec D{(t-1)}$, input into the PatchMorph-Block for the current scale, and new updates $\vec d_{\text{out}_{(t)}}$ are obtained. These updates are then integrated into $\vec D_{(t)}$, with overlapping patch results being averaged.

The required number of patches for completing a scale $t$ is determined by the ratio of an individual patch’s volume to that of the coordinate field, ensuring that each voxel in $\vec D_{t}$ is derived from an average of 10 patch predictions $\vec d_{\text{in}_{(t)}}$.

\subsubsection{CNN Architectures}

In our experiments, we tested PatchMorph based on two kinds of CNN architectures: an affine CNN for local affine alignment, inspired by GlobalNet  \cite{hu2018label}, and a dense CNN that allows for single-voxel displacement, similar to VoxelMorph \cite{balakrishnan2019voxelmorph,dalca2018unsupervised} or LocalNet \cite{hu2018label}.

\begin{figure}
\includegraphics[width = \columnwidth]{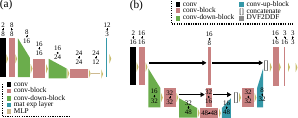}
\caption{Panel (a) shows the CNN architecture used to predict an affine DDF. Panel (b) shows the U-Net architecture that predicts a DDF with voxel-level displacements. Number of input and output channels are depiced on top of each layer.  }
\label{fig:netU}
\end{figure}

Our CNN network architectures are built on four major components: the 3D convolution, a convolution-block, a down-convolution block, and an up-convolution block. The convolution-block consists of a batch norm, followed by a leaky ReLU and a 3D convolution, if not stated otherwise, with a kernel size of 3. A Down-convolution block is identical to a Convolution-block, but with the stride parameter set two 2. Consequently, it halves the spatial resolution of a patch. The up-convolution block is a convolution-block, but with a nearest neighbor up-scaling operation with a scaling factor of 2, placed before the convolution. An instance normalization layer (not shown) is normalizing the input patches. Figure \ref{fig:netU} shows a sketch of our CNN architectures along the input and output channels of each layer.

\smallskip
\textbf{Affine CNN:} Panel (a) in Figure \ref{fig:netU} shows the simple encoder network that predicts 12 parameters of an affine transformation.  This matrix is used to predict the DDF using a matrix exponential layer: Let $\vec A$ be the 12 parameters in the form of an affine matrix. Let $\vec T=e^{(0.25\vec A)}$ be the estimated affine matrix, were $e^{\vec A}$ is the matrix exponential of a square matrix $\vec A$ that we use as activation function. We heuristically determined the constant $0.25$ so that the network generated small deformations with randomly initialized network parameters. Let $\vec T_{3\times 3}$ be the left upper $3\times3$ matrix of $\vec T$, representing rotation and scale, and $\boldsymbol \tau$ be the 3D translation vector of $\vec T$.
Further, let $\vec r$ be a vector from the patch coordinate field which was normalized to the interval $[0,1]$. Then the matrix exponential layer is defined by:
\begin{align}
f(\vec r):=(\vec T_{3\times 3} (\vec r-0.5) + 0.5 + 0.1  \boldsymbol \tau) - \vec r.
\end{align}
The result is then scaled by the patch size to obtain the DDF  $\hat{\vec d}_\text{local}$ in (patch) voxel spacing. Note that for the zero tensor, the matrix exponential is just the identity transform. The translation factor 0.1 has been determined heuristically. In our implementation, we used the approximation $e^{\vec A}\approx (\vec A/k+\vec I)^k$, with $k=10$ for performance reasons. 

\smallskip
\textbf{Dense CNN:} 
We estimate voxel-displacement fields using a simplified U-Net in combination with an integration layer to estimate a DDF from a dense velocity field (DVF), similar to the diffeomorphic VoxelMorph. We used the DVF2DDF block from  MONAI \cite{monai2023}. Panel (b) in Figure \ref{fig:netU} illustrates the architecture.  The last convolution has a kernel size of 1.

\subsubsection{Patches: sizes and  resolution}
In our figures, we show results for nine distinct scales. The number of which was determined through a preliminary experiment (details provided later in the experiments). We determine the coarsest and finest resolutions of the patches based on training data, with intermediate resolutions linearly interpolated between these two extremes. All patches possess isotropic voxel resolutions.

Using a uniform array size for the patches, for instance, $32^3$, we cycle through all training images. During each cycle, we calculate the maximum edge length (in millimeters) across all canvases $\vec I_\text{ref}$, which determines the edge length for the coarsest patch scale. Thus, at the coarsest scale, a patch, when centered on an image, encompasses the entire image in low resolution.

The voxel resolution for the most detailed patch scale is set to match the smallest voxel resolution found along any axis in all training images. Consequently, this voxel resolution is always equal to or finer than that of a source image.

\subsection{Patch-wise Displacement Field Regularization}

Common regularization terms in dense displacement fields enforce diffeomorphic properties to ensure invertible mappings. These terms typically rely on the first- or second-order derivatives of the coordinate field to represent smoothness or bending energy, respectively \cite{rueckert1999nonrigid}. Another approach uses the determinant of the Jacobian matrix to monitor local volume changes \cite{wolterink2022implicit,byra2023exploring}, where a determinant of one indicates volume preservation and values at or below zero suggest non-invertible foldings-outcomes our method aims to prevent.

In PatchMorph, we employ a combination of bending energy \cite{rueckert1999nonrigid} (implementation from MONAI \cite{monai2023}) and a new coordinate field prior to counteracting singularities and foldings. Both regularization terms are derived from the patch coordinate field which was normalized to voxel units, given by $\Phi=\vec X_{\text{moved}_{(t)}}/\vec s$, where $\vec s$ represents the spatial resolution of a voxel within the patch.  

Let $|J_{\Phi}(\vec r)|$ represent the determinant of the Jacobian of the normalized coordinate field $\Phi$ at position $\vec r$. A determinant equal to or less than 0 indicates a singularity or a folding in the field. In these cases, the field becomes non-invertible, a situation we aim to prevent. The new prior is expressed as a hinge loss:
\begin{align}
f(\vec r):=\text{Relu}(-w(|J_{\Phi}(\vec r)|-t)),
\label{eq:regterm}
\end{align}
with  heuristic values $t = 0.5$ and $w=1000$. 

This penalty is calculated for each voxel by selecting the maximum value for each batch item and then averaging across the entire batch dimension. While this approach might seem stringent, its effectiveness is underpinned by three key facets of PatchMorph: its small patch size, which confines the impact of high penalties, a large batch size diluting outlier effects for balanced regularization, and a coarse-to-fine strategy that allows vast displacements while preserving voxel topology. 

\section{Experiments}
Our experimental evaluation comprised two parts. First, we conducted an exhaustive parameter search and performance validation using the MindBoggle dataset with human T1 MRI brain scans. Second, to validate the generalizability of our parameters, we applied the same architecture to a distinct dataset of marmoset brain images acquired through microscopy.

\subsection{Datasets}

\subsubsection{MindBoggle Dataset}

The MindBoggle dataset consists of 101 T1-weighted MRI scans \cite{klein2012101}. The dataset provides labels for 62 cortical regions across both hemispheres. Each dataset contains an image in its native image space and a version aligned to a standard MNI template, both accompanied by an affine world transformation matrix.

For our PatchMorph experiments, we used the original, skull-stripped images. Following the partitioning used by  \cite{byra2023exploring}, we randomly divided the dataset into 73 training, 5 validation, and 20 test images, excluding one due to missing data. The remaining 2 images were used as fixed images for evaluation during validation and testing. They were added to the training set as well. The images were taken from 53 men and 47 women, ranging in age from 19 to 61 years. The images varied in array sizes, with a rounded average dimension of $[212\times 250\times 219]$ voxels with a rounded standard deviation of $[41\times 14\times 46]$ voxels, and presented diverse affine transformations, often involving swapped or mirrored axes. While most images had isotropic voxel resolutions of $1~mm^3$, 21 images featured an anisotropic resolution of $1.2 \times 1 \times 1 ~mm^3$.

\subsubsection{Marmoset Brain Dataset}

We utilized a collection of 52 3D marmoset brain images acquired via serial two-photon microscopy  from \cite{skibbe2023brain} for image registration, focusing on the first color channel that captures auto-fluorescent tissue signals. The dataset includes a population average template used as a fixed image in our experiments, along with cortical and subcortical labels for 36 brain regions. These labels were manually annotated on the template and projected onto individual brains using the ANTs toolkit. For our study, we split the dataset evenly, with 27 images for training, including the template, and 26 for testing. We used the raw, unaligned images with a mean size of approximately $212 \times 257 \times 327$ voxels (standard deviation of $13 \times 16 \times 17$) and an isotropic voxel resolution of $(0.1~mm)^3$.

The dataset presents challenges such as variable background signal quality and corruption from anterograde tracer injections, along with tissue damage and missing coronal sections due to acquisition problems.

\subsubsection{Metrics}
We assessed three metrics: average Dice score (Dice$_{\text{avg}}$), minimum Dice score (Dice$_{\text{min}}$), and the integrity of the deformation field ($|J_{\Phi}|\le 0$). For the latter, we calculated the proportion of negative Jacobian determinants from the normalized coordinate field, aiming for a 0\% rate as the ideal. It is important to note that, in contrast to \cite{byra2023exploring}, we excluded voxels outside the brain volume from this analysis, increasing the sensitivity to any anomalies in the deformation field.

\subsubsection{PatchMorph Architecture}

\begin{table}
\small
\centering
\caption{Test results (mean$\pm$std) on the MindBoggle dataset obtained for PatchMorph $32^3$ for 3,6,9 and 12 scales.}\label{tab:nscales}
\begin{adjustbox}{max width=\columnwidth}
\begin{tabular}{|c|c|c|c|l|}
\hline
Scales &Dice$_{\text{avg}}$ $\uparrow$ & Dice$_{\text{min}}$ $\uparrow$ & $|J_{\Phi}|\le$0 {[}\%{]} $\downarrow$   \\ \hline
3& 0.468$\pm$0.033 & 0.163$\pm$0.078 & 0.000062$\pm$0.000188 \\ \hline
6& 0.535$\pm$0.021 & 0.232$\pm$0.098 & 0.000255$\pm$0.000818 \\ \hline
9& 0.546$\pm$0.020 & 0.262$\pm$0.094 & 0.000012$\pm$0.000060 \\ \hline
12& 0.533$\pm$0.022 & 0.251$\pm$0.100 & 0.000009$\pm$0.000041 \\ \hline
\end{tabular}
\end{adjustbox}
\end{table}

We tested three, six, nine and twelve distinct scales. Preliminary experiments determined that nine scales yielded optimal performance on the MindBoggle validation set. Table \ref{tab:nscales} shows the results on the test set. We also tested several patch sizes on the validation set, with $32^3$ performing best (details discussed later in the text).

Subsequently, we evaluated two PatchMorph configurations with nine scales. The first, our default setup, utilizes a shared Affine CNN for the coarsest scales (0-2), a separate Affine CNN for the intermediate scales (3-5), and a Dense CNN for the finest scales (6-8). The second configuration, referred to as A-PatchMorph, employs an additional Affine CNN in place of the Dense CNN for the finest scales.

\subsubsection{PatchMorph Training}
In each training iteration, we process two pairs of moving and fixed images (four images in total). From each pair, we randomly sample 10 patches for each scale, resulting in a total batch size of 20. The training incorporates two data augmentation steps: First, we apply random affine transformations to the initial crop matrices $\vec T_{\text{p}_0}$, adjusting for rotation (within $\pm 15^\circ$) and scale (between 0.9 and 1.1). Second, we apply another set of random affine transformations to the moving image's transformation matrix $\vec T_\text{m}$, introducing variations in rotation (up to $\pm 25^\circ$), scale (ranging from 0.8 to 1.2), and translation (up to $\pm 2$ cm). Note that $\vec T_{\text{p}_0}$ effects both moving and fixed image. We used the image intensities themselves as a saliency map to sample patches with a higher likelihood from the foreground than from the background, maintaining a ratio of 2:1.

We implemented PatchMorph in PyTorch 2.0.1  (https://pytorch.org/).  An AdamW optimizer \cite{loshchilov2017decoupled} with an initial learning rate of 0.001, and gradient norm clipping \cite{pascanu2013difficulty} with a max norm of 2 is used for optimization. The training regimen begins with 10,000 iterations focusing solely on the coarsest scale $t=0$, after which the second scale is included for another 10,000 iterations. This incremental process continues until all 9 scales have been incorporated. Upon reaching 90,000 iterations, we extend the training with 40,000 additional iterations, employing a scheduler to gradually decrease the learning rate. To enhance efficiency, the loss is calculated only for the last three scales, reducing the frequency of patching operations.

Following the loss used by  \cite{byra2023exploring}, for the MindBoggle dataset, a global NCC loss is combined with a local NCC loss using a kernel size of 9. For the Marmoset brain dataset, we implement a mutual information loss, which is detailed later in the text.

\subsection{Results: MindBoggle Dataset}

\subsubsection{Comparison with other methods}

\begin{figure*}
\includegraphics[width = \textwidth]{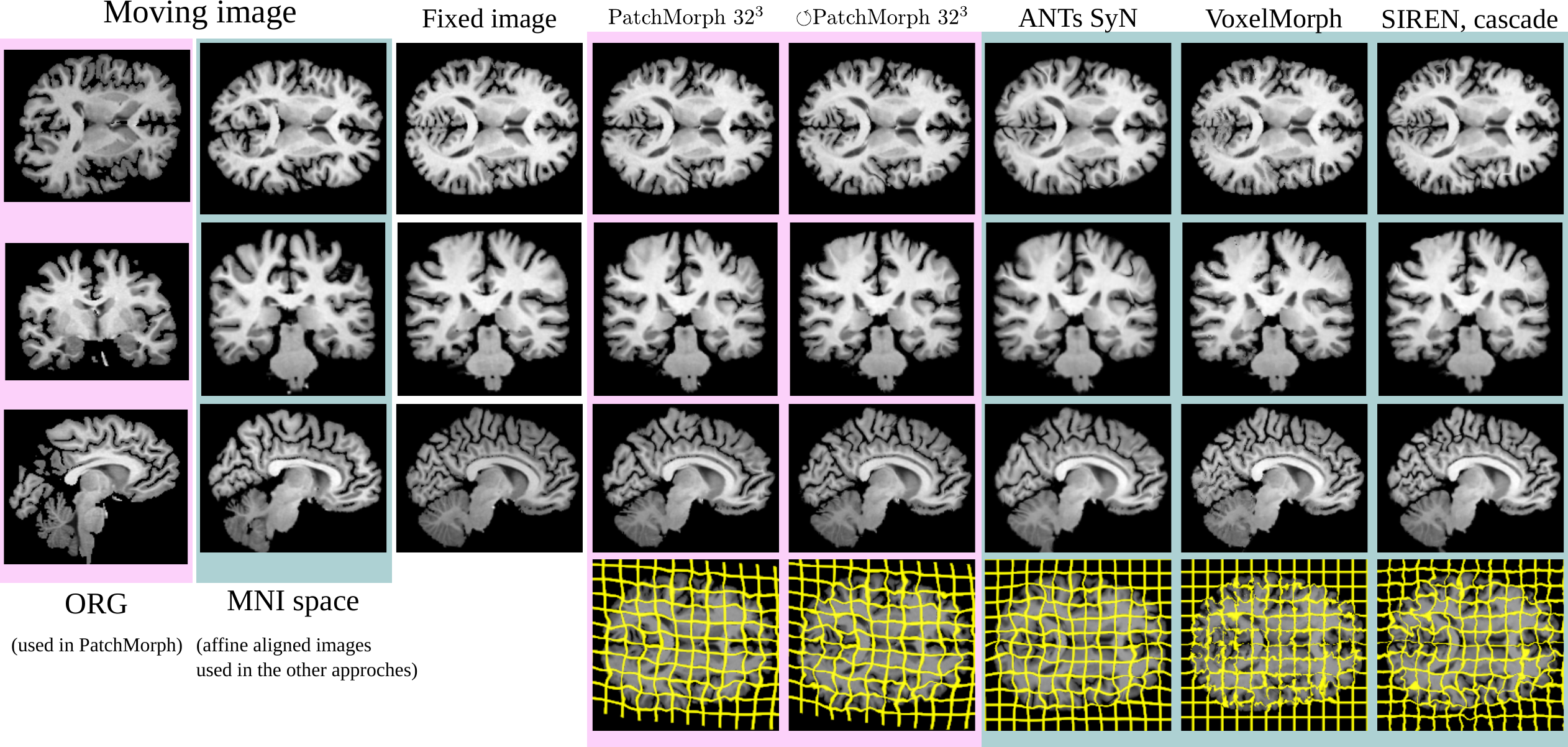}
\caption{A qualitative example from the MindBoggle experiment. For PatchMorph, we used the original image as moving image, while for the other approaches, we used the affine pre-alinged version. PatchMorph and ANTs exhibit the smoothest warp fields, while maintaining good registration results. }
\label{fig:mindbogglewarp}
\end{figure*}

We benchmarked PatchMorph against the methods evaluated in our prior study on implicit neural representation networks (INRs)~\cite{byra2023exploring}. As some methods necessitate pre-aligned images, comparative analyses were performed on affine-aligned brain images. Our comparison included a standard Affine baseline, ANTs SyN registration, diffeomorphic VoxelMorph (D-VMorph), and SIREN, the highest-performing INR model from ~\cite{byra2023exploring}.

The results for PatchMorph are shown in the upper section of Table \ref{tab:extensions_scores}, and the comparative analyses are shown in the lower section. Figure \ref{fig:mindbogglewarp} shows qualitative results for one example.

\begin{table}
\small
\centering
\caption{Test results (mean$\pm$std) on the MindBoggle dataset obtained for the proposed PatchMorph methods.}
\label{tab:extensions_scores}
\begin{adjustbox}{max width=\columnwidth}
\begin{tabular}{|cl|c|c|c|l|}
\hline
ID&Method  & Dice$_{\text{avg}}$ $\uparrow$ & Dice$_{\text{min}}$ $\uparrow$ & $|J_{\Phi}|\le$0 {[}\%{]} $\downarrow$   \\ \hline
\multicolumn{5}{|c|}{Applied to original images  } \\ \hline
1&PatchMorph $16^3$& 0.488$\pm$0.031 & 0.184$\pm$0.091 & 0.003519$\pm$0.006409 \\ \hline
2&PatchMorph $24^3$& 0.533$\pm$0.021 & 0.245$\pm$0.095 & 0.000239$\pm$0.000851 \\ \hline
3&PatchMorph $32^3$& 0.546$\pm$0.020 & 0.262$\pm$0.094 & 0.000012$\pm$0.000060 \\ \hline
4&PatchMorph $48^3$& 0.518$\pm$0.020 & 0.229$\pm$0.086 & $<$0.000001 \\ \hline
5&A-PatchMorph $32^3$& 0.424$\pm$0.019 & 0.193$\pm$0.069 & $<$0.000001 \\ \hline
\multicolumn{5}{|c|}{Applied to original images, without proposed patch regularization  } \\ \hline
6&PatchMorph $32^3$& 0.560$\pm$0.021 & 0.263$\pm$0.094 & 0.446435$\pm$0.216907 \\ \hline
\multicolumn{5}{|c|}{With repetitions (Applied to original images, repeating finest scale twice more) } \\ \hline
7&$\circlearrowleft$PatchMorph $32^3$& 0.583$\pm$0.023 & 0.282$\pm$0.104 & 0.062850$\pm$0.028076 \\ \hline
\multicolumn{5}{c}{  } \\ \hline
\multicolumn{5}{|c|}{Reference (Applied after affine alignment to the templates)} \\ \hline
8& Affine  & 0.325$\pm$0.041 & 0.116$\pm$0.047 & -- \\ \hline
9&{ANTs SyN} & {0.544$\pm$0.019} & {0.260$\pm$0.103} & $<$0.000001  \\ \hline
10&{D-VMorph} & {0.539$\pm$0.131}  & {0.209$\pm$0.077} & {0.655295$\pm$0.053899}   \\ \hline 
11&SIREN & 0.579$\pm$0.112 & 0.260$\pm$0.099 & 2.077100$\pm$0.3304151   \\ \hline 
\end{tabular}
\end{adjustbox}
\end{table}

We tested four patch sizes: $16^3$, $24^3$, $32^3$, and $48^3$ voxels, with $32^3$ yielding the highest Dice score (0.546). In further tests, we repeated certain PatchMorph scales during inference. Considering the vast number of combinations, here we report only the scenario where we repeated the finest scale ($t=8$) on a patch size of $32^3$ twice more.  This approach resulted in the highest Dice score observed (0.583). While SIREN outperformed other methods (0.579), ANTs SyN and PatchMorph showed significantly better deformation field quality, with SIREN struggling in nearly 2\% of voxels. With lower Dice scores (0.424) on the MindBoggle dataset than other methods, A-PatchMorph still significantly outperformed the baseline.

\begin{figure}
\centering
\includegraphics[width = \columnwidth]{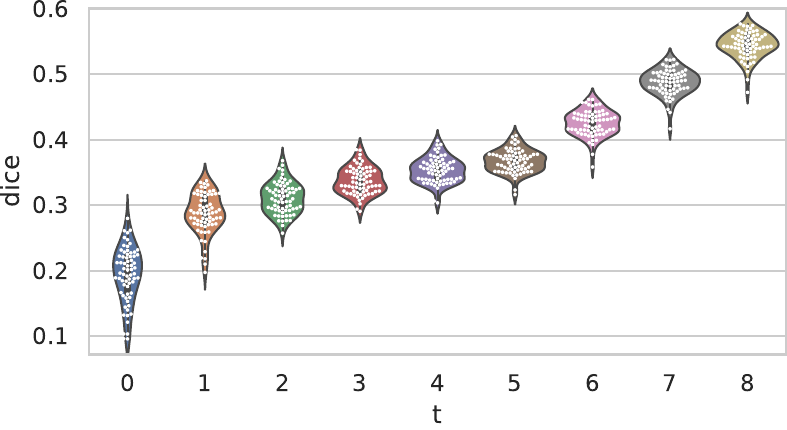}
\caption{Intermediate DICE scores after the first three initial Affine registrations, the intermediate three Affine registrations, and three final deformable registrations of PatchMorph.}
\label{fig:scaleDICE}
\end{figure}

Figure \ref{fig:scaleDICE} shows the dice score assigned to each single scale for PatchMorph $32^3$. It shows that the biggest gain is in the three first and last three scales. 

\subsubsection{Performance Analysis} To comprehensively assess the efficacy and efficiency of PatchMorph, we performed a series of performance analyzes that scrutinize various aspects of the algorithm, ranging from regularization impacts to computational resource utilization.

\smallskip
\noindent
\textbf{Regularization:} Omitting our field regularization in favor of bending energy alone increased Dice scores but at the cost of deformation field quality; see the row with the ID 6 in Table~\ref{tab:extensions_scores}.

\begin{figure}
\centering
\includegraphics[width = \columnwidth]{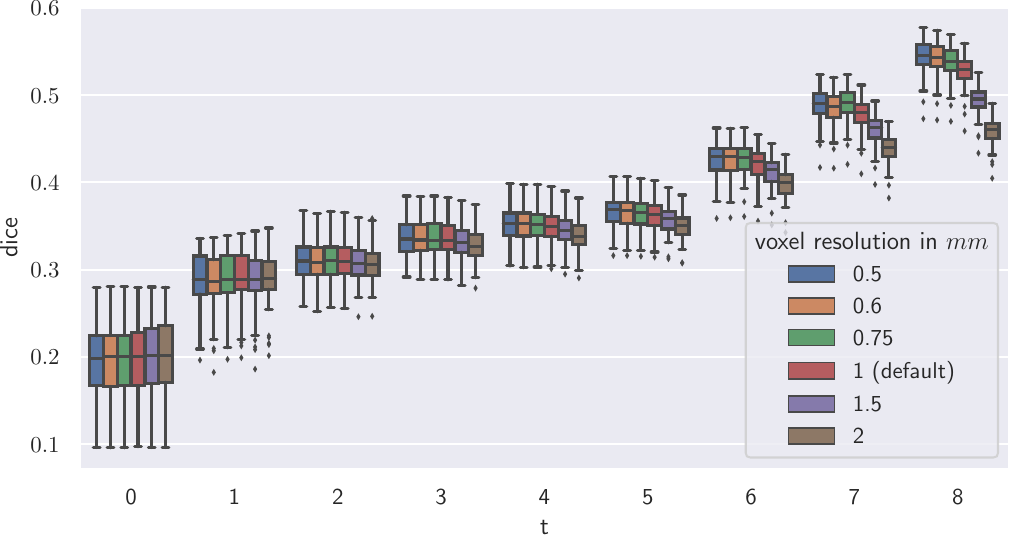}
\caption{Results for PatchMorph $32^3$ with various changes in canvas resolution with respect to the patch resolution. Doubling the canvas resolution lead to the best results, but comes with the costs  that the array size of the resulting warping field will grow accordingly, increasing memory demands.}
\label{fig:canvas_scale}
\end{figure}

\smallskip
\noindent
\textbf{DDF resolution:}  We evaluated various scaling factors for the array resolution of the DDF $\vec D$ in inference. The optimal result was achieved by doubling the baseline resolution from $(1mm)^3$ to $(0.5~mm)^3$. Figure \ref{fig:canvas_scale} illustrates the Dice scores across the range of tested scale factors on the test set.

\smallskip
\noindent
\textbf{Warpfield quality:}
To determine the invertibility of the warp fields, we computed both forward and backward warping fields for each test image using one of the templates. Using ANTs' antsApplyTransforms with linear interpolation, we concatenated both fields to warp a template to an individual dataset and back. Ideally, this results in an identity transform with a Jacobian determinant of 1. As per Table \ref{tab:invertability}, PatchMorph with additional scale repetitions achieved superior Dice scores. However, this also introduced slight errors, leading to a marginally compromised warp field quality.

\begin{table}
\small
\centering
\caption{Diffeomorphic property: Evaluation of the combined forward and backward transformation. Most left number corresponds to the experiment number in Table \ref{tab:extensions_scores} for MindBoggle, and Table \ref{tab:marmoset_extensions_scores} for the Marmoset,respectively.}
\label{tab:invertability}
\begin{adjustbox}{max width=\columnwidth}
\begin{tabular}{|cl|c|c|l|}
\hline
ID&Method  & Median $|J_{\Phi}|$  & $|J_{\Phi}|\le$0 {[}\%{]} $\downarrow$   \\ \hline
\multicolumn{4}{|c|}{MindBoggle Dataset} \\ \hline
3& PatchMorph $32^3$& 0.933$\pm$0.007 & 0.000003$\pm$0.000014 \\ \hline
7& $\circlearrowleft$ PatchMorph $32^3$& 0.871$\pm$0.011 & 0.119557$\pm$0.035596 \\ \hline
\multicolumn{4}{|c|}{Marmoset Dataset} \\ \hline
3& PatchMorph $32^3$& 1.000$\pm$0.014 &  0.003317$\pm$0.005597 \\ \hline
5&  A-PatchMorph $32^3$& 1.005$\pm$0.006 &   0.006845$\pm$0.009445 \\ \hline
\end{tabular}
\end{adjustbox}
\end{table}

\smallskip
\noindent
\textbf{Performance and Operations:} 
Inference with PatchMorph on GPUs was rapid, with times ranging from 2.7 to 7.5 seconds across systems equipped with diverse GPUs, including Nvidia v100, rtxa5000, and h100. Patching was identified as the most memory-intensive operation. In scenarios lacking GPU acceleration, inference times were notably longer, extending to nearly two minutes. Our evaluations also revealed that nearest neighbor interpolation for scattering displacement vectors was not only sufficient but also substantially faster than our custom linear interpolation.

Regarding training requirements, PatchMorph $32^3$ on the MindBoggle dataset with six and nine scales necessitated approximately 5 and 9 gigabytes of GPU memory, respectively. The training duration for models with nine scales varied significantly depending on the system and the load of the system, with times ranging between 7.5 and 20 hours.

\smallskip
\noindent
\textbf{Parameters:} The PatchMorph $32^3$ model has 667,664 trainable parameters, while A-PatchMorph $32^3$ has 558,312.

\subsection{Results: Marmoset Brain Dataset}

\begin{figure}
\includegraphics[width = \columnwidth]{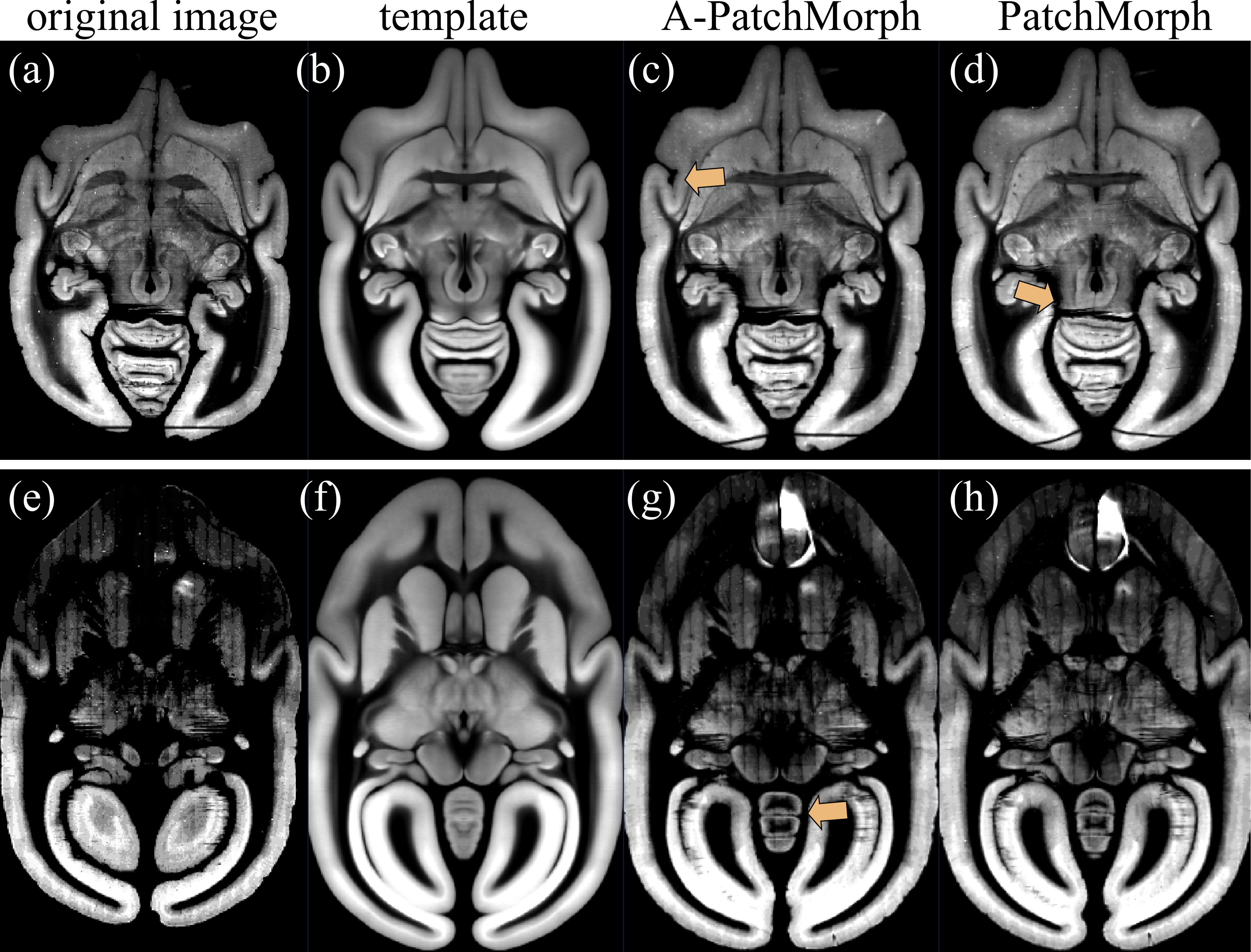}
\caption{
Two challenging examples from the marmoset brain image experiment are depicted here, highlighting issues with missing coronal sections (upper example) and persistent tracer signals (lower example). Each image displays a cross-section from a 3D dataset. The first column shows a section from the original image, followed by a section from the target template in the second column. The last two columns present sections from the moved images using the local affine A-PatchMorph and the PatchMorph architectures, respectively. A-PatchMorph tends to preserve minor irregularities and spatial differences post-alignment (see panels (c) and (g)). In contrast, PatchMorph is more adaptive and can smooth out such irregularities; however, it may also inappropriately close gaps caused by tissue damage, potentially leading to incorrect deformations, as illustrated in panel (d).
}
\label{fig:marmosetexample}
\end{figure}

The Marmoset dataset poses an out-of-distribution challenge, with artifacts not fully represented in the training set. In this experiment, we compared PatchMorph against ANTs results from \cite{skibbe2023brain}, using the same $32^3$ patch size for both PatchMorph and its variant, A-PatchMorph, as in the Mindboggle dataset. To address misalignments due to tracer signal leakage and contrast differences, we employed a mutual information (MI) loss from MONAI during training, replacing the cross-correlation metric. Additionally, we implemented a localized MI loss by dividing patches into $8^3$ sub-cubes and averaging their MI values, each loss weighted equally at 0.5.

The results, presented in Table \ref{tab:marmoset_extensions_scores}, show that both PatchMorph versions achieved comparable Dice scores around 0.89. However, further scrutiny revealed unique challenges: the standard PatchMorph occasionally over-stretched tissues to compensate for damage or missing sections (Fig.~\ref{fig:marmosetexample}, panel (a)), whereas A-PatchMorph's reliance on local affine transformations made it less susceptible to artifacts but also less adept at smoothing irregular tissue boundaries (Fig.~\ref{fig:marmosetexample}, panel (c) and (g)). An invertibility analysis of the warp fields, similar to that conducted for the MindBoggle dataset (refer to Table \ref{tab:invertability}), indicated no significant issues.

\begin{table}
\small
\centering
\caption{Test results (mean$\pm$std) on the Marmoset dataset obtained for the proposed PatchMorph methods.}
\label{tab:marmoset_extensions_scores}
\begin{adjustbox}{max width=\columnwidth}
\begin{tabular}{|cl|c|c|c|l|}
\hline
&Method  & Dice$_{\text{avg}}$ $\uparrow$ & Dice$_{\text{min}}$ $\uparrow$ & $|J_{\Phi}|\le$0 {[}\%{]} $\downarrow$   \\ \hline
5&A-PatchMorph $32^3$& 0.896$\pm$0.015 & 0.716$\pm$0.052 & 0.000005$\pm$0.000028 \\ \hline
3& PatchMorph $32^3$& 0.890$\pm$0.034 & 0.624$\pm$0.133 & 0.000060$\pm$0.000159 \\ \hline
\multicolumn{5}{c}{  } \\ \hline
\multicolumn{5}{|c|}{Reference} \\ \hline
&Translation& 0.423$\pm$0.151 & 0.024$\pm$0.058 & --\\ \hline
&Affine& 0.792$\pm$0.035 & 0.435$\pm$0.121 & -- \\ \hline
\end{tabular}
\end{adjustbox}
\end{table}

\section{Discussion}

In this study, we introduced PatchMorph, a novel image cascade registration approach that operates on compact patches across multiple scales. PatchMorph sequentially refines the focus from coarse to fine details, leveraging stacked CNNs that cascade deformation field updates to enhance registration accuracy. In tests on the MindBoggle dataset, comprising T1-weighted MRI images of human brains, PatchMorph significantly outperformed other traditional and deep learning-based techniques. Further testing on a dataset of microscopy images of marmoset brains demonstrated the approach's generalizability.

We have also demonstrated the effectiveness of the newly proposed coordinate field prior, which proved pivotal for obtaining an invertible transformation between image pairs. Similar to other registration methods such as ANTs \cite{avants2011reproducible} and multiscale versions of VoxelMorph\cite{mok2020large,zhang2021cascaded}, PatchMorph utilizes a multiscale cascade to optimize the registration process in a coarse-to-fine manner. Notably, PatchMorph maintains a consistently compact patch size, which not only minimizes the GPU's memory footprint but also simplifies the handling of world coordinate transformations between two input images, accommodating variances in spacing, array sizes, and orientations.

Our investigation into two types of CNNs for local affine and voxel-based displacements within the PatchMorph framework yielded state-of-the-art results with relatively simple and shallow network architectures. We found that sharing the same network architecture and weights across multiple scales significantly enhances performance without an increase in the number of trainable parameters; see Table~\ref{tab:nscales}.

A potential limitation of PatchMorph is the requirement for patches to form a nested cascade, where the field of view for each patch must not exceed that of its predecessor. This is crucial during training to ensure correct information propagation from one scale to the next. Consequently, it becomes challenging to correct errors made at coarser scales at finer resolutions. 

It is also noteworthy that while PatchMorph is fast during inference, methods like VoxelMorph can be faster. Typically, they are applied once to the entire images, whereas PatchMorph stochastically assembles numerous predictions on small patches. Furthermore, the operations of extracting patches and rendering them back into images are integral to PatchMorph and benefit greatly from fast GPU memory. Currently, running PatchMorph on a CPU is significantly slower than on a GPU and likely slower than approaches akin to VoxelMorph in most scenarios.

Moreover, similar to VoxelMorph, SIREN, or even ANTs, PatchMorph's performance also depends on the choice of loss function. While the prior is crucial for deformation field quality, the metric is central to quantifying image similarity between moved and fixed image patches. Although normalized cross-correlation worked well for the MindBoggle dataset, we had to employ a mutual information loss for the marmoset brain dataset. Yet, corruptions in the data due to leaked neural tracer signals or tissue damage still posed challenges.

The current version of PatchMorph was trained in an unsupervised manner. Supervised training could potentially mitigate difficulties in registering challenging cases, such as leaking tracer signal or missing tissue found within the marmoset brain images. However, creating training data and defining a clear training objective is complex. Related works like SynthMorph \cite{hoffmann2021synthmorph}, which trains on synthesized data, have shown that supervised training does not necessarily outperform unsupervised training. We consider exploring further training strategies, including supervised \cite{eppenhof2018pulmonary} or unsupervised \cite{bigalke2023unsupervised} training approaches in future iterations of PatchMorph.

\section{Conclusion}
PatchMorph introduces a novel multiscale cascade registration approach tailored for brain imaging, adeptly balancing between global and local image attributes. Utilizing compact patches in a cascading refinement process, this method significantly reduces GPU memory usage during the training phase. Our empirical evaluations, conducted on both human and marmoset brain images, demonstrate PatchMorph's versatility across various datasets.

In its current iteration, PatchMorph employs relatively simple CNN architectures. Benefitting from a small memory footprint, PatchMorph opens avenues for the integration of deeper or more sophisticated network structures within this framework, possibilities that were previously untenable with other approaches due to memory limitations.

Moreover, the compact patch size and the stochastic application of PatchMorph hold promise for future iterations. These could potentially include mechanisms to selectively disregard non-registrable sections of an image, such as damaged tissue or lesions, which is particularly pertinent in clinical settings.

To support reproducibility, the source code for PatchMorph will be made publicly available online after acceptance of the paper.

\ifx\ispreprint\undefined
   
\else
    \subsection*{Acknowledgment}
    This work was supported by the program for Brain Mapping by Integrated Neurotechnologies for Disease Studies (Brain/MINDS) from the Japan Agency for Medical Research and Development AMED (JP15dm0207001).
    
\fi

\ifx\ispreprint\undefined
    \bibliographystyle{IEEEtran}
    \bibliography{mybib}
\else
 \bibliographystyle{ieeetr}
 
\footnotesize{

}

\fi

\end{document}